\documentclass[nohyperref]{article}

\usepackage{microtype}
\usepackage{graphicx}
\usepackage{natbib}
\usepackage{subfigure}
\usepackage{booktabs} 

\usepackage{hyperref}

\usepackage{arxiv_template/arxiv}

\usepackage[utf8]{inputenc} 
\usepackage[T1]{fontenc}    
\usepackage{amsmath}
\usepackage{mathtools}
\usepackage{amsthm}
\usepackage{url}            
\usepackage{amsfonts}       
\usepackage{nicefrac}       
\usepackage{lipsum}
\usepackage{amssymb}
\usepackage{listings}       
\usepackage{xcolor}         
\usepackage{svg}            
\usepackage{multirow}       

\definecolor{codegreen}{rgb}{0,0.6,0}
\definecolor{codegray}{rgb}{0.5,0.5,0.5}
\definecolor{codepurple}{rgb}{0.58,0,0.82}
\definecolor{backcolour}{rgb}{0.95,0.95,0.96}

\lstdefinestyle{codestyle}{
    backgroundcolor=\color{backcolour},   
    commentstyle=\color{codegreen},
    keywordstyle=\color{magenta},
    numberstyle=\tiny\color{codegray},
    stringstyle=\color{codepurple},
    basicstyle=\ttfamily\footnotesize,
    breakatwhitespace=false,         
    breaklines=true,                 
    captionpos=b,                    
    keepspaces=true,                 
    numbers=left,                    
    numbersep=5pt,                  
    showspaces=false,                
    showstringspaces=false,
    showtabs=false,                  
    tabsize=2
}
\hypersetup{
    colorlinks=true,
    linkcolor=blue,
    filecolor=magenta,      
    citecolor=blue,
    urlcolor=blue,
    pdftitle={Flashlight - Enabling Innovation in Tools for Machine Learning},
    pdfpagemode=FullScreen,
    }

\newcommand\scalecomponents{1}
\newcommand\scalecomputationmodes{1}

\newcommand\paperreprocode{\href{https://github.com/flashlight/paper}{this URL}}

\title{Flashlight: Enabling Innovation in Tools for Machine Learning}

\author{%
Jacob Kahn \\
Facebook AI Research \\ Menlo Park, CA \\
\texttt{jacobkahn@fb.com} \\
\And
Vineel Pratap \\
Facebook AI Research \\ Menlo Park, CA \\
\texttt{vineelkpratap@fb.com} \\
\And 
Tatiana Likhomanenko \\
Facebook AI Research\thanks{Currently at Apple.} \\ Menlo Park, CA \\
\texttt{antares@fb.com} \\
\And 
Qiantong Xu \\
Facebook AI Research\thanks{Currently at SambaNova Systems.} \\ Menlo Park, CA \\
\texttt{qiantong@fb.com} \\
\And
Awni Hannun \\
Zoom AI \\ San Jose, CA \\
\texttt{awni.hannun@zoom.us} \\
\And
Jeff Cai \\
Facebook AI Research\thanks{Currently independent.} \\ Menlo Park, CA \\
\texttt{jcai@fb.com} \\
\And
Paden Tomasello \\
Facebook AI Research \\ Menlo Park, CA \\
\texttt{padentomasello@fb.com} \\
\And
Ann Lee \\
Facebook AI Research \\ New York, NY \\
\texttt{annl@fb.com} \\
\And
Edouard Grave \\
Facebook AI Research \\ Paris \\
\texttt{egrabve@fb.com} \\
\And
Gilad Avidov \\
Facebook \\ Menlo Park, CA \\
\texttt{avidov@fb.com} \\
\And
Benoit Steiner \\
Facebook AI Research \\ Menlo Park, CA \\
\texttt{benoitsteiner@fb.com} \\
\And
Vitaliy Liptchinsky \\
Facebook AI Research \\ Menlo Park, CA \\
\texttt{vitaliy888@fb.com} \\
\And
Gabriel Synnaeve \\
Facebook AI Research \\ Paris \\
\texttt{gab@fb.com} \\
\And
Ronan Collobert \\
Facebook AI Research\thanks{Currently at Apple.} \\ Menlo Park, CA \\
\texttt{locronan@fb.com} \\
}

\begin{document}
\maketitle

\begin{abstract}

As the computational requirements for machine learning systems and the size and complexity of machine learning frameworks increases, essential framework innovation has become challenging. While computational needs have driven recent compiler, networking, and hardware advancements, utilization of those advancements by machine learning tools is occurring at a slower pace. This is in part due to the difficulties involved in prototyping new computational paradigms with existing frameworks. Large frameworks prioritize machine learning researchers and practitioners as end users and pay comparatively little attention to systems researchers who can push frameworks forward --- we argue that both are equally important stakeholders. We introduce Flashlight, an open-source library built to spur innovation in machine learning tools and systems by prioritizing open, modular, customizable internals and state-of-the-art, research-ready models and training setups across a variety of domains. Flashlight allows systems researchers to rapidly prototype and experiment with novel ideas in machine learning computation and has low overhead, competing with and often outperforming other popular machine learning frameworks. We see Flashlight as a tool enabling research that can benefit widely used libraries downstream and bring machine learning and systems researchers closer together. Flashlight is available at \href{https://github.com/flashlight/flashlight}{this URL}.

\end{abstract}

\keywords{machine learning \and deep learning \and systems, frameworks \and autograd library \and tensor library}

\section{Introduction}
\label{sec:intro}

The recent rise of deep learning-based techniques has been accompanied and sustained by the wide availability of dedicated frameworks such as TensorFlow \citep{abadi2016tensorflow} and PyTorch \citep{paszke2019pytorch}. These frameworks have enabled the democratization of machine learning research by providing extensive collections of high level primitives to support common use cases. Lowering the barrier to entry for end users has boosted the popularity of both neural networks and the frameworks in which they are implemented. However, in order to support what are now vast ecosystems and a diverse user base, framework size and complexity have increased dramatically over time. As a result, deep, groundbreaking framework research has become onerous and time consuming, precluding rapid innovation. Given these barriers, major deep learning frameworks have become entrenched in their existing operating modes.

Innovation in this area remains as important as ever. Indeed, framework innovation accelerates machine learning (ML) and artificial intelligence (AI) research. Frameworks that are easier to use reduce the engineering burden on researchers, and frameworks that are higher-performance decrease the time required to iterate on experimental work and validate hypotheses. Even more critically, tooling plays a fundamental role in deciding which ideas succeed or fail.
For example,~\citet{lecun1989} pioneered the use of convolutional neural networks (CNNs) \citep{fukushima1982455} trained using backpropagation 
for computer vision tasks in the late 1980s, which was subsequently applied to handwriting recognition. However, widespread success for CNNs was achieved two decades later when \citet{alexnet} leveraged the CUDA programming model to take advantage of graphics processing units (GPUs) to train a much deeper model (AlexNet).


While deep learning frameworks have been optimized to leverage existing hardware paradigms for common neural network architectures, they often fail to deliver similar efficiencies on designs that diverge from the mainstream. For example, \citet{ml_systems_stuck} explain how the design of these frameworks results in poor hardware utilization for a novel type of neural network, known as a capsule network \citep{46653}, that leverages new components such as squashing operations and routing by agreement. More generally, what are now unconventional approaches to modern problems in machine learning require highly-specialized additions to popular frameworks. As a result of narrowly-optimized systems, research beyond deep learning may be discounted due to purported computational infeasibility given modern frameworks' capabilities.

Furthermore, the waning of Moore's law~\citep{7878935} coupled with the ever-growing computational demands of deep learning are prompting several shifts in hardware. Massive-scale distributed computing is now required to train leading models --- a process that established frameworks remain unable to handle truly automatically --- that is, without some manual specification as to how to distribute work. In parallel, multiple specialized hardware products are now available to better support deep learning applications: Nvidia's TensorCores \citep{NVIDIATC}, Google's TPUs \citep{tpu}, Graphcore's IPUs \citep{IPU}, Apple's Neural Engine\footnote{\url{https://nr.apple.com/dE9q1p9M7t}}, and others have been developed to improve total float-pointing operations (FLOPs), cost per-FLOP, or energy consumption. Additionally, numerous efforts are underway to move away from conventional von Neumann computing architectures in which memory and processing units are physically separated, either by storing data closer to compute units or by switching to in-memory computing altogether.

While tooling innovation is alive and well given these incentives for progress, working within large, well-established frameworks has become more and more challenging as framework size and scope grows. While some tools such as Halide \citep{halide_autoscheduler, MLSYS2021_73278a4a} and TVM \citep{tvm, ansor} are built from first-principles, many recent innovations have required the development of ad-hoc tools. For example, FlexFlow \citep{flexflow, roc} underpins recent work aimed at improving the use of distributed computing to accelerate the training of large neural networks; PET \citep{pet} provides a framework that enables graph-level neural network optimizations; and DeepSpeed \citep{rasley2020deepspeed} implements algorithms supporting custom distribution of computation. With ad-hoc approaches, researchers are required to start from scratch for new directions or adapt their ideas to fit into the scaffolding these frameworks provide --- resulting in significant technical burdens.

To sustain framework innovation, we introduce Flashlight, an open source minimalist ML library designed to support research in machine learning frameworks, facilitate rapid iteration on ideas, reduce the engineering burden on researchers, and remove the need for new tools. Flashlight includes:
\begin{itemize}
  \item A modular, component-based architecture that makes every aspect of the implementation fully \textbf{customizable} with simple internal APIs.
  \item A compact yet highly-performant \textbf{reference implementation} of each component.
  \item A comprehensive set of \textbf{benchmarks} representative of the state-of-the-art in machine learning on which to evaluate alternative implementations.
\end{itemize}




\section{Related Work}
\label{sec:background}

Numerous frameworks have been implemented in recent years to support machine learning, including Lush \citep{bottou2002lush}, Theano \citep{bergstra2010theano}, Torch \citep{torch7}, Caffe \citep{jia2014caffe}, MXNet \citep{chen2015mxnet}, deeplearning4j \citep{dl4j}, TensorFlow \citep{abadi2016tensorflow}, Flux \citep{innes2018flux}, Jax \citep{jax2018github}, PyTorch \citep{paszke2019pytorch},  Chainer \citep{tokui2019chainer}, and PaddlePaddle \citep{ma2019paddlepaddle}. These frameworks offer programming models designed around multidimensional arrays (\textsc{Tensors}), modeled as first-class objects and supported by a comprehensive set of mathematical primitives (or operators) to manipulate them. To provide the computing power required by deep learning-based methods, most natively support hardware accelerators such as general-purpose GPUs or custom-designed ASICs such as TPUs.

Generally, framework implementations follow one of a few computational models:
\begin{itemize}
    \item In the \textit{deferred execution} model, the neural network to be trained is first encoded as a dataflow graph which can be optimized for a specific set of target hardware devices. The neural network is then executed in a distinct second phase. Since the dataflow graph represents the entire computation, both local and global optimizations can be applied, making the subsequent execution very efficient. However, only programs that can be represented as dataflow graphs can be processed with this approach, thus limiting flexibility. Frameworks such as Theano, Caffe, TensorFlow 2.0, or MXNet fall into this category.

    \item In the \textit{eager} model, an interpreter (such as Python) is extended with the high level kernel-based operations needed to train a neural network. These operations are executed immediately when called, though this precludes many optimizations. By weaving neural network-related operations into a Turing complete programming language, this approach is extremely flexible. Furthermore, the imperative nature of the underlying programming language allows for fine-grained control over the execution order and memory utilization, which enables more specific user-driven optimization. Frameworks such as Torch, TensorFlow 2.0 Eager, PyTorch, or Chainer exemplify this approach.
    
    \item In the \textit{static} model, computation is explicitly specified ahead-of-time either via an implicit or explicit schedule. Operations are executed inside runtime sessions. Given that the entire graph of computation is fully-specified before execution, significant global optimizations can be applied here, such as explicit ahead-of-time (AOT) scheduling. Frameworks such as TensorFlow 1.0 fall into this category.
    
    \item The \textit{hybrid} model simply combines multiple of the above computation models.
\end{itemize}






\section{Principles}
\label{sec:flgoals}

The aforementioned frameworks are designed and implemented to best-serve their user bases --- namely, machine learning researchers and practitioners. They rely on large, internally complex codebases to provide comprehensive solutions, as is further discussed in Section~\ref{sec:results}.

In contrast, Flashlight targets an audience of researchers interested in experimenting with new designs and implementations of machine learning tools or broader computational or modeling paradigms. To foster this type of innovation, Flashlight balances simplicity and nimbleness with the need to provide enough functionality to support real use cases. Internal and external simplicity is the key design principle of Flashlight; the ability to dramatically modify software and drive it in new directions is inversely correlated with codebase size and complexity \citep{gill}. More specifically:
\begin{itemize}
\item Flashlight is built on a shallow stack of \textbf{idiomatic, modular, and customizable} abstractions. Framework components interact through small, well-defined, stable APIs, which expose most internal aspects of its implementation. This ensures that every component of Flashlight can be modified or replaced with new custom implementations, even e.g. its memory manager and tensor implementation. To support the exploration of a wide array of alternative approaches, Flashlight interfaces are \textbf{flexible and unopiniated} by design. This is in contrast to other frameworks, which impose stricter implementation requirements based on tight design constraints for their computation models and support requirements across hardware, downstream frameworks, and other ecosystem members.

\item Flashlight provides deliberately-\textbf{compact default implementations} of its APIs. This reduces out-of-the-gate engineering burden and the need for modifications, and enables fast compilation and rapid iteration when experimenting. Furthermore, to mitigate premature optimization, Flashlight deliberately \textbf{abstains from adding small efficiency improvements} if they conflict with the goals of keeping the codebase simple and APIs clean.

\item Flashlight is a \textbf{research-first} framework, and is not intended for out of the box production use. To keep codebase size small, it forgoes features such as model servers for deployment and integration with cluster management tools.

\end{itemize}

Flashlight is a viable solution for \textbf{machine learning research}, shipping with a comprehensive set of benchmarks and research setups for state-of-the-art neural network architectures such as convolutional neural networks (CNNs) \citep{krizhevsky12} and Transformers \citep{transformers}, as well as task-specific models such as ViT \citep{dosovitskiy2020image}, DETR \citep{carion2020end}, or BERT \citep{devlin2018bert}. The speech recognition system wav2letter \citep{wav2letter++}, is also built entirely on Flashlight.

Benchmarks built on these state-of-the-art models make Flashlight a \textbf{turn key solution for system researchers} who want to quickly evaluate their design and implementation choices without needing to build test benches from the ground-up. More importantly, Flashlight makes possible end-to-end benchmarking on real models rather than microbenchmarks or small-scale tests.

\section{Design}
\label{sec:fldesign}

\begin{figure*}[t]
\begin{center}
    \includegraphics[width=\scalecomponents\columnwidth]{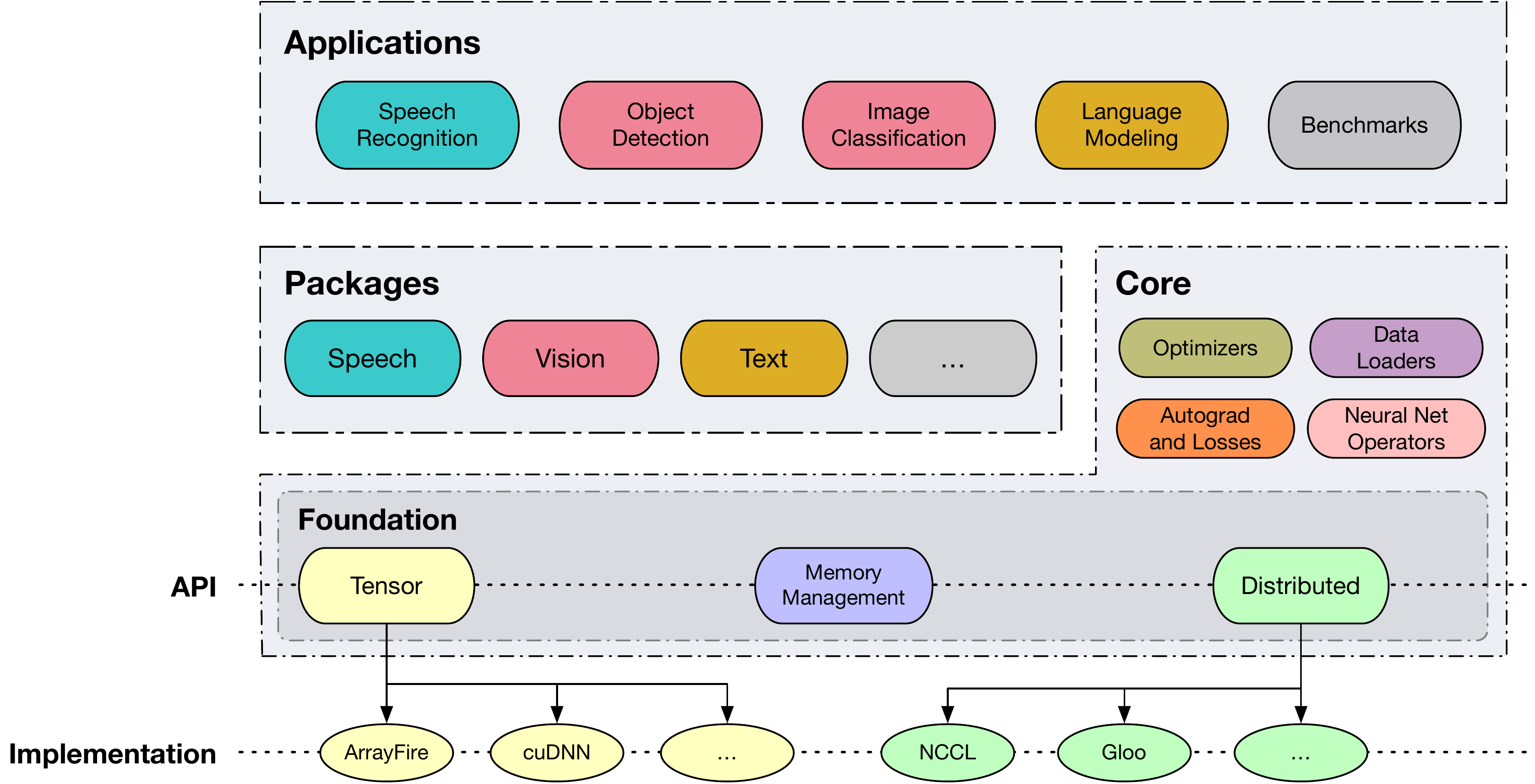}
\end{center}
\caption{Components of the Flashlight library.}
\label{fig:components}
\end{figure*}

Flashlight's design is centered around \textit{internal} APIs for framework components which form the building blocks for domain-specific ML \textit{packages} and \textit{applications} --- this structure is outlined in Figure \ref{fig:components}. Flashlight is implemented as a C++ library and follows a Tensor-based programming methodology, with neural network building blocks that derive from a \textsc{Module} interface, communicate by exchanging Tensor data, and are composed functionally or imperatively to form complete neural network architectures. Tensor programming in Flashlight is fundamentally dynamic, but given that C++ is a compiled language, code describing models in Flashlight is compiled. This approach promotes type safety, foregoes the runtime overheads associated with interpreters, and, unlike eager-based approaches, enables global optimizations where possible.

\subsection{Open Foundational Interfaces}
\label{sec:fl_design_fundation}

Flashlight is built on top of three open \textit{foundational} APIs, each addressing design and implementation challenges faced by machine and deep learning tools: a \textit{Tensor} interface, a \textit{memory management} subsystem, and a \textit{distributed} computing interface. These APIs are backed by reference implementations that enable Flashlight to efficiently target CPUs, GPUs, and other accelerators. These include code generation and dedicated kernels for Intel, AMD, OpenCL, and CUDA devices, and leverage libraries such as cuDNN \citep{chetlur2014cudnn}, 
MKL \citep{mkl}, oneDNN \citep{onednn}, ArrayFire \citep{Yalamanchili2015}, and MiOpen \citep{khan2019miopen}.

\subsubsection{Tensor Interface}
\label{sec:tensor_interface}

Modern deep learning frameworks feature tensor library internals which sit under deep layers of abstractions, requiring numerous framework modifications in order to iterate on tensor stack design. Flashlight's \textsc{Tensor} abstraction is defined in terms of existing tensor libraries via a simple, extensible interface and a high-level API that mirrors \textit{numpy} \citep{harris2020array} rather than using specific, opinionated intermediate representations (IRs) or large operator sets.

\begin{figure*}[t]
\begin{center}
    \includegraphics[width=\scalecomputationmodes\columnwidth]{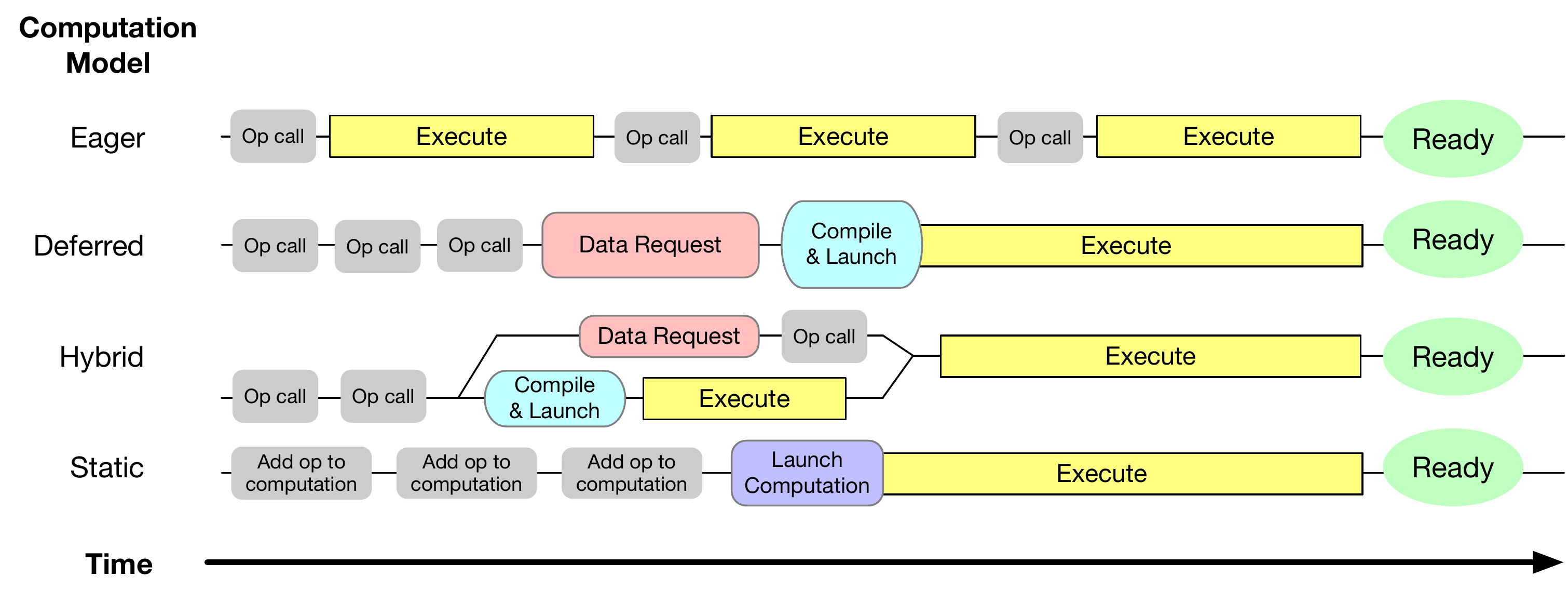}
\end{center}
\caption{Flashlight's Tensor API supports backend implementations with any of the above computation modes (or entirely new modes that may result from further research).}
\label{fig:computation_models}
\end{figure*}

Flashlight \textsc{Tensor} backend implementations need not follow any particular computation mode as outlined in Section~\ref{sec:background} and shown in Figure~\ref{fig:computation_models}. Tensor values need only be materialized upon user request --- typically when extracting the output values of a model or inspecting intermediary state. This provides a flexibility unique amongst deep learning frameworks to either defer or eagerly-compute intermediate values --- or to experiment with new computation paradigms altogether.

Implementing a \textsc{Tensor} backend in Flashlight involves fulfilling a small set of implementation requirements. Users have full control of their implementations after subclassing two interfaces:
\begin{itemize}
    \item the \textsc{TensorAdapter} interface (Listing \ref{lst:tensor_interface}) which allows a backend to attach custom stateful information and metadata to each tensor. This includes shape, type, and memory information~--- which may be implementation-dependent.
    \item the \textsc{TensorBackend} interface (Listing \ref{lst:tensor_backend}) which allows backends to store global state as needed (e.g device compute streams, dataflow graphs) and implement a small set of primitive tensor operations including unary and binary operations (e.g. arithmetic ops), reductions, matrix multiplication, and convolution.
\end{itemize}

Rather than require implementations of large, highly-specialized operator sets or interoperability with complex dispatch mechanisms or intermediate representations (IRs) as do other frameworks, Flashlight operators outside of the small \textsc{TensorBackend} API are derived by composition. For example: the ReLU activation is implemented by leveraging the \textsc{max} operator. Flashlight's reference \textsc{Tensor} implementation uses a \textit{hybrid} approach, offloading computation to highly-optimized vendor libraries when advantageous and relying on deferred, on-the-fly code generation via ArrayFire for all other operations so as to increase kernel arithmetic intensity.



\begin{minipage}{\linewidth}
\begin{lstlisting}[language={C++}, numbers=none, caption={The \textit{TensorAdapter} interface for implementing operations on tensor metadata and storing tensor state for individual tensor instances.}, label={lst:tensor_interface}]
class MyTensorImpl : public TensorAdapter {
  // State information goes here
  // (e.g. buffers, shape)
 public:
  // Metadata
  const Shape& shape() override;
  dtype type() override;
  // ...
};
\end{lstlisting}
\end{minipage}

\begin{minipage}{\linewidth}
\begin{lstlisting}[language={C++}, numbers=none, caption={The \textit{TensorBackend} interface for implementing operations on tensors and storing global backend state.}, label={lst:tensor_backend}]
class MyTensorBackend : public TensorBackend {
  // State information goes here (e.g. compute streams, compiler state)
 public:
  // Tensor operation primitives
  Tensor add(const Tensor& lhs, const Tensor& rhs) override;
  Tensor minimum(const Tensor& lhs, const Tensor& rhs) override;
  // ...
};
\end{lstlisting}
\end{minipage}

\subsubsection{Memory Management}
\label{sec:design:memory}
Robust memory management is an important research area as model size increases. While individual \textsc{Tensor} backends in Flashlight can perform their own memory management as defined by implementers, Flashlight's default \textsc{Tensor} backend also provides a generic API for defining custom memory management schemes. By default, memory is only allocated when needed for just-in-time compilation. A sample of this API is shown in Listing \ref{lst:code_samples:memory}. To support the lazy computation model as well as just in time code generation, memory allocations only occur when tensors need to be materialized per the compute graph. Buffers are used asynchronously after they are requested depending on the timing of kernel launches but are not freed until computation is complete.


\subsubsection{Distributed Training}
\label{sec:design:distributed}
Flashlight provides a low-level API for distributed training primitives with performant default implementations in GPU and CPU settings using \citet{nccl} and \citet{gloo}, respectively. Users can add new backends or custom methods of performing distributed computation, and can use primitives in other framework components as needed. Further details and samples can be found in Section~\ref{sec:app:code_samples:distributed_training}.

\begin{lstlisting}[language={C++}, numbers=none, caption={An implementation of a memory manager using the memory management API.}, label={lst:code_samples:memory}]
class CachingMemoryManager : public MemoryManagerAdapter {
  // Store state as needed
 public:
  void* alloc(bool userLock, unsigned ndim,
    dim_t* dims, unsigned elSize) override;
  void unlock(void* ptr, bool userLock) override; // free memory
  // ...
};
\end{lstlisting}

\subsection{The Flashlight Core}
\label{sec:design:core}
The \textsc{Tensor} API, together with memory management and distributed computation abstractions provide a foundation on which to build other core machine learning algorithms and applications. These other core components are outlined below. Section~\ref{sec:app:code_samples} provides code samples and linking documentation for the below components.

\paragraph{Neural Network Primitives} To facilitate the implementation of neural networks, Flashlight ships with common neural building-blocks encompassing activation functions, normalization layers, regularizers, losses, and more. These derive from the \textsc{Module} abstraction, as discussed above, which provides a method of chaining and nesting operations. Section~\ref{sec:app:code_samples:modules} contains more details and sample implementations.

\paragraph{Automatic Differentiation} Automatic differentiation (autograd) is implemented via a simple \textsc{Variable} abstraction. A Variable takes a \textsc{Tensor} argument when created, and its underlying Tensor (or gradient Tensor) can be accessed at any time. Variables feature operators which call underlying Tensor operations and record those operations to a dynamic tape in a design similar to \citet{Paszke2017AutomaticDI} while being lightweight enough to allow implementations of other autograd paradigms. In keeping with Flashlight's modularity, \textsc{Tensor} and \textsc{Variable} are separated to avoid performance and implementation overhead in non-gradient-based ML algorithms. Further, this gives implementers of a Flashlight Tensor backend an efficient, fully-featured autograd system with no additional effort. We provide an example of an autograd primitive implementation in Listing \ref{lst:code_sample_autograd}.


\paragraph{Optimizers} Flashlight provides implementations of most common first-order stochastic optimization algorithms, as included in other frameworks. These are defined in terms of Variable and Tensor operations, allowing for open-ended experimentation (e.g. with distributed computation, in-place operations, etc).

\paragraph{Data Loaders} Flashlight provides a simple \textsc{Dataset} class which abstracts away the notion of a \emph{sample} in ML algorithms. A sample is viewed here as a \textsc{Tensor} or vector of \textsc{Tensors}. Datasets are trivially composable to create pipelines to transform, resample, or parallelize (via native C++ threads) the construction of such samples. While Flashlight core dataset abstractions are agnostic from the end-user task, data-specific datasets are provided in higher-level \textit{packages}, to efficiently load from disk structured data (e.g. images, audio or text).

\begin{lstlisting}[float=*t, language={C++}, numbers=none, caption={Defining a cosine autograd operator in Flashlight using \textsc{Tensor} operations and \textsc{Variable}.}, label={lst:code_sample_autograd}]
Variable cos(const Variable& input) {
  auto result = fl::cos(input.tensor()); // get a Tensor from a Variable
  // Called with backward() to compute gradients for this op's inputs
  auto gradFunc = [](std::vector<Variable>& inputs,
                     const Variable& gradOutput) {
    inputs[0].addGrad( // Add a gradient to the input
        Variable(gradOutput * negate(sin(inputs[0].tensor())), false));
  };
  // Construct a Variable from a Tensor and a gradient-computing function
  return Variable(result, {input}, gradFunc);
}
\end{lstlisting}

\subsection{Packages and Applications}
\label{sec:packages_and_applications}
Flashlight contains additional domain-specific abstractions leveraging both core components as well as stand-alone implementations. These abstractions allow end-users or ML researchers to quickly get started on various ML applications. The \textit{package} module provides building blocks for common ML tasks, domain-specific algorithms, and helpers. The \textit{application} module leverages these building blocks to provide complete, ready to use solutions (e.g. models, training loops, evaluation pipelines). When not original to Flashlight, implementations reproduce the task performance those they reference. We leverage several of these applications to evaluate Flashlight's performance in Section~\ref{sec:results}.

\paragraph{Speech.} Flashlight provides an implementation of classical featurization (spectogram, log-mel filterbanks, etc.) that can run on-the-fly with minimal overhead. It also provides a collection of data-augmentation techniques (including additive noise and reverberation), as well as implementations of speech-specific sequential criteria and model architectures. Flashlight contains a fast beam-search decoder (which can interface any language model) and beam rescorers \citep{wav2letter,wav2letter++}. Research performed with the speech \textit{application} have reached and are competitive with state-of-the-art results \citep{synnaeve2019end,likhomanenko2020rethinking}.

\paragraph{Vision.} Flashlight offers built-in data loaders for standard computer vision benchmarks (such as ImageNet \citep{imagenet} and COCO \citep{coco}) along with large set of efficient data-augmentations and transformations. It includes mainstream image classification models: convolutional (e.g. ResNet \cite{he2016deep}) and transformer-based architectures (e.g. ViT \cite{dosovitskiy2020image}), as well as a modern, transformer-based object detection model (DETR \cite{carion2020end}) and helpers (e.g. Hungarian matching and object detection evaluation). 

\paragraph{Text.} Flashlight ships with support for text dataset manipulation and tokenization along with language modeling training pipelines for a variety of neural language models, including transformer \citep{transformers} and CNN-based \citep{dauphin2017gated}. Both autoregressive and masked, e.g. BERT, language modeling tasks are supported. These language models can be combined with other domain-specific packages such as speech.

\section{Evaluation}
\label{sec:results}

In the sections that follow, we compare Flashlight to two widely-used deep learning frameworks --- PyTorch and Tensorflow --- with metrics relevant to framework research velocity. We also evaluate framework performance to demonstrate the potential of our approach and the quality of the default implementations of all our components. We outline the steps needed to reproduce all our results in the Appendix.

\subsection{Code Complexity}
\label{sec:code_complexity}

Flashlight is built to minimize complexity and operating surface. As frameworks grow and are combined with other frameworks or take on new platform-specific requirements, internal modifiability decreases. Table~\ref{table:complexity} compares frameworks across well-established measures of complexity and portability --- binary size, lines of code, and number of operators and operator implementations. Flashlight's small surface facilitates easily exploring new designs and prototyping on new hardware --- having few sources of truth simplifies the process of replacing core components and ensures end-to-end tests don't opaquely fall back to existing implementations.

\begin{table*}[t]
\caption{Complexity of various frameworks based on high-level metrics. We provide additional analysis disambiguating tensor library components of each framework in Section \ref{sec:app:code_complexity:lib_size}.}
\label{table:complexity}
\vskip 0.15in
\begin{center}
\begin{small}
\begin{sc}
\begin{tabular}{c c c c c}
\toprule
Metric & PyTorch & TensorFlow & (Ours) Flashlight \\
\midrule
Binary Size (MB) & 527 & 768 & 10 \\
Lines of Code & 1,798,292 & 1,306,159 & 27,173 \\
Number of Operators & 2,166 & 1,423 & 60 \\
\midrule
\textbf{Approx num. ops. that perform:} \\
\textsc{add}  & 55 & 20 & 1  \\
\textsc{conv} & 85 & 30 & 2  \\
\textsc{sum}  & 25 & 10 & 1  \\
\bottomrule
\end{tabular}
\end{sc}
\end{small}
\end{center}
\vskip -0.1in
\end{table*}

\subsubsection{Compilation Time}
\label{sec:compilation_time}

When modifying or adding significant new research code to framework internals, recompilation can be costly. Large frameworks depend on code generation for broad platform support\footnote{Examples | PyTorch: \url{https://git.io/Jzel9}, TensorFlow: \url{https://git.io/JzeRw}}, increasing compilation time. Further, expensive incremental rebuilds can slow iteration speed.

Flashlight is sufficiently-lightweight and modular so as to enable from-source build times that are orders of magnitude faster than other frameworks, as shown in Table~\ref{table:compile_times}. Times were measured for both from-scratch and incremental builds with Intel Xeon Gold 6138 CPUs with 80 cores and 750 GB of memory. To estimate incremental build performance, we randomly sample 100 source files without replacement, make trivial modifications that force recompilation, and time the resulting rebuild. While we constrain the set of files that can be modified for incremental compilation to those that are part of core systems (tensor library, autograd, modules, distributed computation), framework subsystems differ and constraints were specialised per-framework. The standard deviation for all incremental compilation benchmarks was under 5\% of the overall compilation time. The Appendix contains information and resources to reproduce these results.

\begin{table}[t]
\caption{Compile times in CPU minutes across frameworks, with incremental compilation times averaged over 100 samples. Flashlight compilation time includes compiling with the default ArrayFire CUDA backend. }
\label{table:compile_times}
\vskip 0.15in
\begin{center}
\begin{small}
\begin{sc}
\begin{tabular}{c c c}
\toprule
Platform & From Scratch & Incremental \\
\midrule

PyTorch & 754 & 132 \\
Tensorflow & 2061 & 371 \\
(Ours) Flashlight & 34 & 0.6 \\

\bottomrule
\end{tabular}
\end{sc}
\end{small}
\end{center}
\vskip -0.1in
\end{table}

\subsubsection{Performance}
\label{sec:performance}



\begin{table*}[ht]
\caption{Performance on common state-of-the-art models across frameworks. Values are the number of seconds needed to perform 100 iterations of the forward and backwards passes, with data loading (unless indicated). Number of parameters are in millions. Framework labels: PT = PyTorch, TF = TensorFlow, and FL = Flashlight.}
\label{table:performance}
\vskip 0.15in
\begin{center}
\begin{small}
\begin{sc}
\setlength{\tabcolsep}{5pt}

\begin{tabular}{lrrrrrrrr}
\toprule
\multicolumn{1}{c}{\multirow{2}{*}{Model}} & & & \multicolumn{3}{c}{1 GPU} & \multicolumn{3}{c}{8 GPUs}       
\\ \cmidrule{4-6}  \cmidrule{7-9}
\multicolumn{1}{c}{} & \multicolumn{1}{c}{Num. Params (M)}  & \multicolumn{1}{c}{Batch Size} & \multicolumn{1}{c}{PT} & \multicolumn{1}{c}{TF} & \multicolumn{1}{c}{FL} & \multicolumn{1}{c}{PT} & \multicolumn{1}{c}{TF} & \multicolumn{1}{c}{FL} \\
\midrule
AlexNet     & 61  & 32  &  2.0   & 4.0    & \bf 1.4   & 6.0   & 6.5   & \bf 2.1  \\
VGG16       & 138 & 32  & 14.8  & \bf 12.6   & 13.2  & 16.3  & 17.9  & \bf 14.9   \\
ResNet-50   & 25  & 32  & 11.1  & 12.4   & \bf 10.3  & 12.3  & 15.9  & \bf 11.9  \\
BERT-like   & 406 & 128 & 19.6  & 19.8   & \bf 17.5  & 22.7  & 23.6  & \bf 19.2   \\
ASR Tr.     & 263 & 10  & 58.5  & 63.7   & \bf 53.6  & 63.7  & 69.7  & \bf 57.5   \\
ViT         & 87  & 128  & 137.8 & 140.3  & \bf 129.3 & 143.1 & 169.6 & \bf 141.0 \\
\bottomrule
\end{tabular}
\end{sc}
\end{small}
\end{center}
\vskip -0.1in
\end{table*}

When improving framework components or modifying internals, framework overhead makes it difficult to disambiguate performance changes due to in-flight modifications from existing bottlenecks or overhead due to other framework components as discussed in Section~\ref{sec:flgoals}. Table~\ref{table:performance} compares the performance of Flashlight 0.3.1, PyTorch 1.8, TensorFlow 2.4 on six common large-scale deep neural networks. For each configuration, we benchmark 100 iterations of data loading\footnote{To ensure fairness, due to Flashlight's significantly better dataloading performance as to compared to other frameworks, BERT-like models use random data in-memory; ViT models exclude data augmentation.}, preprocessing, and forward/backward passes, with data-parallel gradient synchronization in distributed settings. Benchmarks are performed on Intel E5-2698 CPUs with 512GB of RAM, and NVIDIA V100-32GB GPUs in a DGX-1 server. Inter-GPU interconnects in the 8 GPUs (1 node) setting are Nvidia NVLink-based. All models were warmed up with 100 iterations of forward and backward passes. For consistency and reproducibility, no third-party libraries are used to enhance performance beyond optimization tools already contained in frameworks (e.g. \texttt{@tf.function} in TensorFlow). Flashlight is benchmarked as is with no optimizations. While orthogonal to the paper, Flashlight's default backend has empirically outperformed other frameworks due to the quality of the ArrayFire JIT, dataloading performance, and low framework overhead.

Flashlight is competitive and can exceed the performance of other frameworks, especially on architectures which are of lower arithmetic intensity and spend less compute time in vendor-optimized libraries, such as AlexNet. Given strong performance with simple reference implementations that have undergone far less optimization than have large frameworks, we see exciting potential for improvement with future research done in Flashlight.

\subsection{Case Studies}

Ongoing research efforts enabled by Flashlight include work in code generation, compilers and IRs, memory management, and distributed computing. Below, we give examples of recent research made possible with Flashlight.

\subsubsection{Optimizations on Large, Specialized Autograd Graphs}
The ability to change the lightweight implementation of Flashlight's tensor and automatic differentiation (autograd) components via extensible APIs facilitated research in building a fully differentiable beam search decoder \citep{collobert2019fully}, which required operating on unconventional computation graphs not supported by other frameworks' autograd systems. Other frameworks were unable to handle these autograd graphs for several reasons:
\begin{itemize}
    \item Graphs contained millions of nodes/operations that created significant memory pressure;
    \item There existed only small operator overhead per autograd graph node (many addition and $\log{}$ operations);
    \item Graph operations had few opportunities for vectorization;
    \item Only sparse components of the graph were required.
\end{itemize}

Authors modified Flashlight's autograd to support:
\begin{itemize}
    \item On-the-fly graph pruning to take advantage of sparsity and reduce memory footprint;
    \item Dynamic, pre-fused gradient computation for common sequences of gradient computation operations;
    \item Custom autograd node lifetime for avoiding reference-counting overhead for graph mutations.
\end{itemize}

To our knowledge, these capabilities only exist in autograd implementations like Flashlight's that feature public APIs built for customization. Indeed, these ideas may be broadly applicable to other modeling settings, but implementing and researching such logic in other frameworks' autograd systems is not realistically possible.

\subsubsection{Fragmentation Reduction in Accelerator Memory Management}
While those researching memory management techniques in machine learning computation can make ad-hoc modifications to other frameworks' memory managers, build time, internal complexity, and lack of a unified interface makes this challenging. Using Flashlight's open memory management interface, researchers studied and developed new techniques for fragmentation reduction for memory management on the GPU. This was made possible by the ease of extending a lightweight memory manager interface in Flashlight along with clear implementation requirements and tests that made rapid prototyping possible.

Various caching memory allocators are used across deep learning frameworks to reduce the cost of native device memory allocations and reuse already-allocated memory. These caching memory managers are subject to fragmentation as they bucket allocations based on rounded size. Reducing this fragmentation allows for training larger models and significantly improves performance as it removes sometimes-expensive allocation overhead.

Given these challenges, researchers aiming to reduce external fragmentation implemented a custom caching memory manager in Flashlight to study memory behavior and built highly-specialized telemetry that tied individual tensor operations to specific allocations. Researchers detailed a myriad of prototypes to reduce fragmentation and described rapid rebuilds and custom telemetry as critical to their experimentation. Ultimately, a memory manager that restricted splitting large cache blocks (or blocks beyond a certain tunable size) showed promise and reduced internal fragmentation for most models by over 20\%.

These techniques were tested on a variety of models in Flashlight before researchers shared their findings with maintainers of other large deep learning frameworks, where knowledge were contributed upstream\footnote{\href{https://github.com/pytorch/pytorch/pull/44742}{Caching allocator improvements in PyTorch.}}.


\subsubsection{Automatic Optimization of Memory and Distributed Compute}

Flashlight enables developing generalized approaches to both memory management and distributed computation. Approaches such as ZeRO \citep{rajbhandari2020zero} optimizer state sharding or GPUSwap \citep{kehne2015gpuswap} memory oversubscription, for example, are specialized to particular components of training pipelines. Generalizations of these approaches which involve shuffling around buffers to a variety of devices or performing pieces of computation on certain hardware should discover new techniques for efficient large-scale training.

Given that Flashlight offers complete control over memory and distributed computation models, tensors can follow any preordained allocation schedule or rules. They can be sharded or computations dispatched to arbitrary devices. Operating under such general assumptions in other frameworks that are without internal APIs for memory management or feature strict requirements around computation model makes specifying such a general system difficult.

\subsubsection{Researching Efficient Primitives in Tensor Computation}

New ideas in basic tensor computation are notoriously difficult to test at scale in an end-to-end fashion. For demonstrative purposes, consider a research artifact which includes new, more-efficient ways to perform element-wise arithmetic operations on tensors (e.g. addition, multiplication). While such operations can be implemented as one-off operators in most machine learning frameworks, swapping out the source of truth for the tensor \texttt{add} function, for instance, such that existing operators and framework components (including baselines, benchmarks, etc) uses the custom implementation may be difficult or untenable. Across various frameworks, this process might involve:
\begin{itemize}
    \item \textbf{PyTorch} --- per Figure~\ref{table:complexity}, 55 PyTorch operators explicitly perform addition as part of their computation (and perhaps far more implicitly); to benchmark end-to-end with existing PyTorch code, all of these callsites must be changed to call the new implementation.
    \item \textbf{TensorFlow} --- similarly to PyTorch, many operators perform addition, and while defining custom operators is straightforwards, changing a wealth of existing library code is error-prone and labor-intensive.
    \item \textbf{Jax} --- defining a custom operator in Jax and using it via composition with other operators is relatively simple. However, swapping out default behavior of addition for most all Jax operators would require changes to XLA via MLIR, a large, complex codebase.
    \item \textbf{Flashlight} --- given the default backend, an implementer can simply subclass or swap out the existing implementation of the \texttt{add} function with their custom logic. All \texttt{add} operations in Flashlight dispatch to that operator, so existing baselines and operations will run with the new implementation without any additional code changes.
\end{itemize}
\section{Conclusion}
\label{sec:conclusion}

We presented Flashlight, a modular machine learning library supporting modern, state-of-the-art baselines that features orders of magnitude less code and binary size as compared to frameworks such as PyTorch and TensorFlow. These large frameworks come fully-featured; Flashlight aims to complement them in providing a first-of-its-kind tool with which to do machine learning \textit{framework} and \textit{computational} research. To this end, Flashlight features a lightweight, modular design, as well as full implementations of mainstream models across a variety of domains, making it easy for researchers to implement new internal tensor, memory management, or distributed computation backends. Flashlight includes a variety of reference implementations for its APIs that compete with and often outperform popular machine learning frameworks, thus demonstrating the viability of our approach.

\subsubsection*{Acknowledgements}

We would like to thank Chaitanya Talnikar and Mohammed Motamedi for valuable contributions to the Flashlight codebase, Shubho Sengupta for valuable support and input, as well as Horace He and Edward Yang for their work on operator size and complexity in deep learning frameworks.


\bibliography{references}
\bibliographystyle{unsrtnat}

\newpage
\appendix

\section{Appendix}
\subsection{Reproducibility}
\label{sec:reproducibility}

All tools and code used in our evaluation are included as supplementary material --- this includes scripts to reproduce benchmarks and other quantitative codebase metrics. Flashlight code can be found on Github at \paperreprocode.


As discussed in Section~\ref{sec:results}, Flashlight v0.3.1 is used to reproduce results using ArrayFire 3.8 as the underlying tensor backend. No other specialized configuration is used for either Flashlight or PyTorch or TensorFlow.

\subsubsection{From-Source Compilation}
\label{sec:app:repro_from_source_compilation}
Flashlight is built in CMake release mode, which is also the default for both PyTorch and TensorFlow builds. Build settings are kept default for all frameworks; Flashlight uses Ninja\footnote{\url{https://ninja-build.org/}} with CMake in accordance with PyTorch's build, while Tensorflow employs Bazel.

Exact build step reproduction can be found in the aforementioned repository.

\subsubsection{Incremental Compilation}
\label{sec:app:incremental_compilation}

Incremental compilation benchmarks are performed using similar build setups per Section~\ref{sec:app:repro_from_source_compilation}. To test incremental rebuilds, source files are randomly selected without replacement from a distribution constructing by weighting each file by the number of lines in the file, and using that number to determine the probability of selecting it for modification.

Scripts to perform and time this incremental compilation can be found in the aforementioned public repository.

\subsection{Code Complexity}
\subsubsection{Operator Counting}
\label{sec:app:code_complexity:operator_counting}
To count the number of operators for each framework, we utilize operator schemas for PyTorch and Tensorflow (which generate code from those schemas, accordingly) written in YAML and Protobuf, respectively. For Flashlight, we count the number of functions in the Flashlight \textsc{Tensor} interface and autograd interfaces, as these form the full implementation requirements for a full tensor backend that functions on all platforms. The scripts released on Github detail the files and filtering techniques used to reproduce the number of results.

To count the number of operators for each implementation, we use the above operator lists, then count the number of operators that perform the specified function, even if those operators perform other functions. For example: an operator called \textsc{addmm}, which performs an addition operation followed by a matrix-matrix multiplication, performs an addition operation, and would thus be counted when tallying the number of \textsc{add} operators.

\subsubsection{Codebase Complexity Disambiguating Tensor Library Size}
\label{sec:app:code_complexity:lib_size}

Given that Flashlight can be compiled with arbitrary tensor libraries that are adapted to the Flashlight Tensor framework, we provide brief further evaluation here exhibiting the contribution of tensor libraries to overall framework size. In an effort to remote auxiliary and irrelevant framework components, both this analysis and the analysis from Section~\ref{sec:results} counts only C, C++, Python, YAML, CUDA, and CMake files from a relevant subset of core framework components when assessing overall lines of code.

Table~\ref{table:complexity_size_detailed} details the relative sizes and number of lines per framework with and without tensor libraries. Note that PyTorch cannot be compiled without its tensor components, so one cannot directly assess binary size.

\begin{table*}[ht]
\caption{Complexity of various framework components including and excluding building with tensor library components.}
\label{table:complexity_size_detailed}
\vskip 0.15in
\begin{center}
\begin{small}
\begin{sc}
\begin{tabular}{c c c c c}
\toprule
Metric & PyTorch & TensorFlow & (Ours) Flashlight \\
\midrule
Binary Size (MB) (no tensor lib) & N/A & 423 & 7 \\
Lines of Code (no tensor lib) & 924k & 602k & 27k \\
Binary Size (MB) (with tensor lib) & 527 & 768 & 10 \\
Lines of Code (with tensor lib) & 1798k & 1306k & 17k \\
Number of Operators & 2,166 & 1,423 & 60 \\
\bottomrule
\end{tabular}
\end{sc}
\end{small}
\end{center}
\vskip -0.1in
\end{table*}

\subsection{Performance}

The aforementioned public repository provides scripts required to fully-reproduce all performance measures.

\subsection{Design Details and Code Samples}
\label{sec:app:code_samples}
In the following sections, we show brief code samples expounding on those in Section~\ref{sec:fldesign}.

\subsubsection{Distributed Training}
\label{sec:app:code_samples:distributed_training}

Flashlight's distributed training API is of a similar flavor to its Tensor library, in that it invites custom implementations of distributed computation primitives with an explicit API. The API is unopinionated and supports both synchronous and asynchronous communication, unlike other frameworks. Alsi included is an internal API for implementing specialized rendezvous schemes for new distributed computation environments. Listing \ref{lst:app:code_sample_distributed_api} shows part of this API, which is structured in a similar manner to \citep{li2020pytorch}. Implementers need simply derive from this interface, and their distributed computation primitives will interoperate with other Flashlight computation streams.

\begin{lstlisting}[language={C++}, numbers=none, caption={Part of Flashlight's distributed computation API.}, label={lst:app:code_sample_distributed_api}]
class DistributedInterface {
  // store arbitrary state here
 public:
  // metadata about the distributed computation environment
  virtual int getWorldRank() = 0;
  virtual int getWorldSize() = 0;
  // ...
  
  // distributed computation primitives
  virtual void allReduce(Tensor& var, double scale = 1.0, bool async = false) = 0;
  virtual void allReduceMultiple(
    std::vector<Tensor> vars,
    double scale = 1.0,
    bool async = false) = 0;
  // ...
    
  // synchronization primitives
  virtual void syncDistributed() = 0;
  virtual void barrier() = 0;
  // ...
};
\end{lstlisting}

\subsubsection{Modules}
\label{sec:app:code_samples:modules}

Flashlight's \textsc{Module} abstraction is similar to that of frameworks such as Torch and PyTorch. It can recursively store other modules and interoperate with more sophisticated abstractions including \textsc{Container}, which wraps multiple modules, \textsc{Sequential}, which stores sequences of modules and forwards data through them sequentially, and user-defined abstractions. Listing \ref{lst:app:code_samples:mnist:model} in Section~\ref{sec:app:code_samples:mnist} shows an example of Sequential usage.

Listing \ref{lst:app:code_sample_modules} shows a small Dropout module implementation that calls into the dropout autograd primitive, stores and serializes a small amount of state, and defines a simple forward function on a Variable.

\begin{lstlisting}[language={C++}, numbers=none, caption={A Dropout layer implemented as a Flashlight module.}, label={lst:app:code_sample_modules}]
class Dropout : public Module {
 private:
  double ratio_;
  FL_SAVE_LOAD_WITH_BASE(Module, ratio_) // serialization

 public:
  Dropout(double drop_ratio = 0.5);
  Variable forward(const Variable& input) override {
    if (train_) {
      return dropout(input, ratio_); // autograd primitive
    } else {
      return input;
    }
  }
  // ...
};
\end{lstlisting}

The \texttt{FL\_SAVE\_LOAD\_WITH\_BASE} macro defines serialization of the Dropout class as a module, including any fields to be serialized (in this case, only the dropout ratio).

\subsubsection{An End-to-End Example: MNIST}
\label{sec:app:code_samples:mnist}

Below, we detail a simple end-to-end training setup following Flashlight's open-source documentation\footnote{\url{https://fl.readthedocs.io/en/latest/mnist.html}}.

First, data is loaded using the \textsc{BatchDataset} abstraction in Listing \ref{lst:app:code_samples:mnist:dataset}:

\begin{lstlisting}[language={C++}, numbers=none, caption={Loading MNIST data into a train and evaluation set.}, label={lst:app:code_samples:mnist:dataset}]
const int kTrainSize = 60000;
const int kValSize = 5000;

auto& [train_x, train_y] = load_dataset(data_dir);

// Hold out a dev set
auto val_x = train_x(span, span, range(0, kValSize));
train_x = train_x(span, span, range(kValSize, kTrainSize));
auto val_y = train_y(range(0, kValSize));
train_y = train_y(range(kValSize, kTrainSize));

// Make the training batch dataset
BatchDataset trainset(
    std::make_shared<TensorDataset>(std::vector<Tensor>{train_x, train_y}),
    batch_size);

// Make the validation batch dataset
BatchDataset valset(
    std::make_shared<TensorDataset>(std::vector<Tensor>{val_x, val_y}),
    batch_size);
\end{lstlisting}

A full description of the \texttt{load\_dataset} function can be found in the MNIST training example on Github\footnote{\url{https://git.io/JVI6O}}.

We can construct the model using a simple \textsc{Sequential} in Listing \ref{lst:app:code_samples:mnist:model}:

\begin{lstlisting}[language={C++}, numbers=none, caption={Constructing a CNN for MNIST training.}, label={lst:app:code_samples:mnist:model}]
const int kImageDim = 28;
auto pad = PaddingMode::SAME;

Sequential model;
model.add(View({kImageDim, kImageDim, 1, -1})); // WHCN (col major)
model.add(Conv2D(
    1 /* input channels */,
    32 /* output channels */,
    5 /* kernel width */,
    5 /* kernel height */,
    1 /* stride x */,
    1 /* stride y */,
    pad /* padding mode */,
    pad /* padding mode */));
model.add(ReLU());
model.add(Pool2D(
    2 /* kernel width */,
    2 /* kernel height */,
    2 /* stride x */,
    2 /* stride y */));
    
model.add(Conv2D(32, 64, 5, 5, 1, 1, pad, pad));
model.add(ReLU());
model.add(Pool2D(2, 2, 2, 2));
model.add(View({7 * 7 * 64, -1}));
model.add(Linear(7 * 7 * 64, 1024));
model.add(ReLU());
model.add(Dropout(0.5));
model.add(Linear(1024, 10));
model.add(LogSoftmax());
\end{lstlisting}

In Listing \ref{lst:app:code_samples:mnist:train}, we create a simple custom training loop. This uses optimizer, loss function, and meter abstractions as provided by default by Flashlight. We perform the forward and backward pass, step the optimizer to update parameters, and zero out gradients before moving to the next batch. We evaluate the model using the function defined in Listing \ref{lst:app:code_samples:mnist:eval}, pulls out the max prediction and comparing it against the ground truth, updating the loss meter as we go, then returning the final loss values.

\begin{lstlisting}[language={C++}, numbers=none, caption={A simple training loop.}, label={lst:app:code_samples:mnist:train}]
// Make the optimizer
SGDOptimizer opt(model.params(), learning_rate);

// The main training loop
for (int e = 0; e < epochs; e++) {
  AverageValueMeter train_loss_meter;

  // Get an iterator over the data
  for (auto& example : dataset) {
    auto inputs = noGrad(example[INPUT_IDX]);
    auto output = model(inputs);

    auto target = noGrad(example[TARGET_IDX]);
    // Compute and record the loss.
    auto loss = categoricalCrossEntropy(output, target);
    train_loss_meter.add(loss.tensor().scalar<float>());
    // Backprop, update the weights and then zero the gradients.
    loss.backward();
    opt.step();
    opt.zeroGrad();
  }

  double train_loss = train_loss_meter.value()(0).scalar<double>();

  // Evaluate on the dev set.
  auto [val_loss, val_error] = eval_loop(model, valset);

  std::cout << "Epoch " << e << std::setprecision(3)
            << ": Avg Train Loss: " << train_loss
            << " Validation Loss: " << val_loss
            << " Validation Error (%): " << val_error << std::endl;
}
\end{lstlisting}

\begin{lstlisting}[language={C++}, numbers=none, caption={Evaluating a training model on MNIST.}, label={lst:app:code_samples:mnist:eval}]
std::pair<double, double> eval_loop(Sequential& model, BatchDataset& dataset) {
  AverageValueMeter loss_meter;
  FrameErrorMeter error_meter;

  // Place the model in eval mode.
  model.eval();
  for (auto& example : dataset) {
    auto inputs = noGrad(example[INPUT_IDX]);
    auto output = model(inputs);
    // Get the predictions in max_ids
    Tensor max_vals, max_ids;
    max(max_vals, max_ids, output.tensor(), 0);
    
    auto target = noGrad(example[TARGET_IDX]);
    // Compute and record the prediction error.
    error_meter.add(reorder(max_ids, 1, 0), target.tensor());
    // Compute and record the loss.
    auto loss = categoricalCrossEntropy(output, target);
    loss_meter.add(loss.tensor().scalar<float>());
  }
  // Place the model back into train mode.
  model.train();

  double error = error_meter.value().scalar<double>();
  double loss = loss_meter.value().scalar<double>();
  return std::make_pair(loss, error);
}
\end{lstlisting}

Finally, in Listing \ref{lst:app:code_samples:mnist:test}, we evaluate the trained model on the test set by creating a test dataset and using the previously-defined evaluation function.

\begin{lstlisting}[language={C++}, numbers=none, caption={Evaluating a training model on MNIST.}, label={lst:app:code_samples:mnist:test}]
std::pair<double, double> eval_loop(Sequential& model, BatchDataset& dataset) {
  AverageValueMeter loss_meter;
  FrameErrorMeter error_meter;

  // Place the model in eval mode.
  model.eval();
  for (auto& example : dataset) {
    auto inputs = noGrad(example[INPUT_IDX]);
    auto output = model(inputs);
    // Get the predictions in max_ids
    Tensor max_vals, max_ids;
    max(max_vals, max_ids, output.tensor(), 0);
    
    auto target = noGrad(example[TARGET_IDX]);
    // Compute and record the prediction error.
    error_meter.add(reorder(max_ids, 1, 0), target.tensor());
    // Compute and record the loss.
    auto loss = categoricalCrossEntropy(output, target);
    loss_meter.add(loss.tensor().scalar<float>());
  }
  // Place the model back into train mode.
  model.train();

  double error = error_meter.value().scalar<double>();
  double loss = loss_meter.value().scalar<double>();
  return std::make_pair(loss, error);
}
\end{lstlisting}

\end{document}